\documentclass[]{gmaart2022}

\renewcommand{\varProcName}{Proc. 35. Workshop Computational Intelligence}
\renewcommand{\varLocation}{Berlin}
\renewcommand{\varDate}{20.-21.11.2025}

\usepackage[utf8]{inputenc} 
\usepackage[T1]{fontenc} 

\usepackage{amsmath,amssymb,amsfonts,mathtools,amsthm} 

\usepackage[babel,autostyle]{csquotes}

\usepackage{tabularx}
\newcolumntype{L}[1]{>{\raggedright\arraybackslash}p{#1}} 
\newcolumntype{C}[1]{>{\centering\arraybackslash}p{#1}} 
\newcolumntype{R}[1]{>{\raggedleft\arraybackslash}p{#1}} 
\usepackage{booktabs}

\usepackage{paralist}   

\usepackage[noend]{algpseudocode}

\algrenewcommand\algorithmicrequire{\textbf{Voraussetzung:}}
\algrenewcommand\algorithmicensure{\textbf{Abschlussbedingung:}}

\usepackage{listings} 
\usepackage{subcaption}  

\usepackage{epstopdf}
\usepackage{xcolor}
\selectcolormodel{gray}  

\usepackage{scrhack}
\usepackage{varwidth}
\usepackage{rotating} 
\usepackage[defaultlines=2,all]{nowidow}  
\usepackage{import}
\usepackage{placeins}
\usepackage{makecell}


\usepackage{amsmath,amssymb,amsfonts,mathtools,amsthm} 

\usepackage{longtable} 
\usepackage{multirow}
\usepackage{tikz}
\usepackage{pgfplots}
\usepackage{smartdiagram}
\usetikzlibrary{patterns}
\usetikzlibrary{shapes} 
\usetikzlibrary{shapes.geometric, arrows,positioning}
\usepackage{smartdiagram}
\usesmartdiagramlibrary{additions} 
\AtBeginDocument{
}
\usepackage{xcolor}
\selectcolormodel{gray}

\usepackage[acronym]{glossaries} 
\usepackage{hyphenat}
\usepackage{etoolbox}
\usepackage{printlen}
\usepackage{graphicx}
\usetikzlibrary{calc, positioning, matrix}
\usepackage{standalone}

\newcommand{\newacr}[4][]{\newacronym[
	sort={\ifthenelse{\isempty{#1}}{#2}{#1}},
	]{#2}{#3}{#4}}
\glsdisablehyper

\robustify{\bfseries}

\begin{document}


\hyphenpenalty=2000

\pagenumbering{roman}
\cleardoublepage
\setcounter{page}{1}
\pagestyle{scrheadings}
\pagenumbering{arabic}

\setnowidow[2]
\setnoclub[2]

\renewcommand{\Title}{Embedding World Knowledge into Tabular Models: Towards Best Practices for Embedding Pipeline Design}

\renewcommand{\Authors}{Oksana Kolomenko\textsuperscript{*}, Ricardo Knauer\textsuperscript{*}, Erik Rodner}
\renewcommand{\Affiliations}{
KI-Werkstatt, University of Applied Sciences Berlin\\
HTW Berlin, 10313 Berlin, Germany\\
E-Mail: ricardo.knauer@htw-berlin.de\\[1ex]
\textsuperscript{*}\textit{These authors contributed equally.}}

\renewcommand{\AuthorsTOC}{O. Kolomenko, R. Knauer, E. Rodner}
\renewcommand{\AffiliationsTOC}{KI-Werkstatt, University of Applied Sciences Berlin}

\setLanguageEnglish				 
\setupPaper

\section*{Abstract}

Embeddings are a powerful way to enrich data-driven machine learning models with the world knowledge of large language models (LLMs). Yet, there is limited evidence on how to design effective LLM-based embedding pipelines for tabular prediction. In this work, we systematically benchmark 256 pipeline configurations, covering 8 preprocessing strategies, 16 embedding models, and 2 downstream models. Our results show that it strongly depends on the specific pipeline design whether incorporating the prior knowledge of LLMs improves the predictive performance. In general, concatenating embeddings tends to outperform replacing the original columns with embeddings. Larger embedding models tend to yield better results, while public leaderboard rankings and model popularity are poor performance indicators. Finally, gradient boosting decision trees tend to be strong downstream models. Our findings provide researchers and practitioners with guidance for building more effective embedding pipelines for tabular prediction tasks.

\section{Introduction}

Large language models (LLMs), pretrained on massive web-crawled corpora, possess a rich world knowledge that can not only be leveraged for tasks in natural language processing, but also in tabular prediction \cite{Knauer2025,Peters2018}. Tabular rows are commonly serialized into text and then converted into numerical representations (embeddings) suitable for downstream models \cite{Carballo2023,Fang2024}. This allows for efficiently augmenting data-driven machine learning methods with LLMs' compressed knowledge about the world.

Despite the growing interest in this paradigm \cite{Grinsztajn2023,Spinaci2024,Kasneci2024,Kim2024,Koloski2025}, there is limited evidence on how to design effective embedding pipelines for tabular prediction, particularly regarding the preprocessing strategies, the selection of embedding models, and the choice of downstream models. This makes it difficult both to generalize insights from academic research and to develop effective embedding pipelines in practice.

\begin{figure}[tbp]
	\centering
	\includegraphics[width=\textwidth]{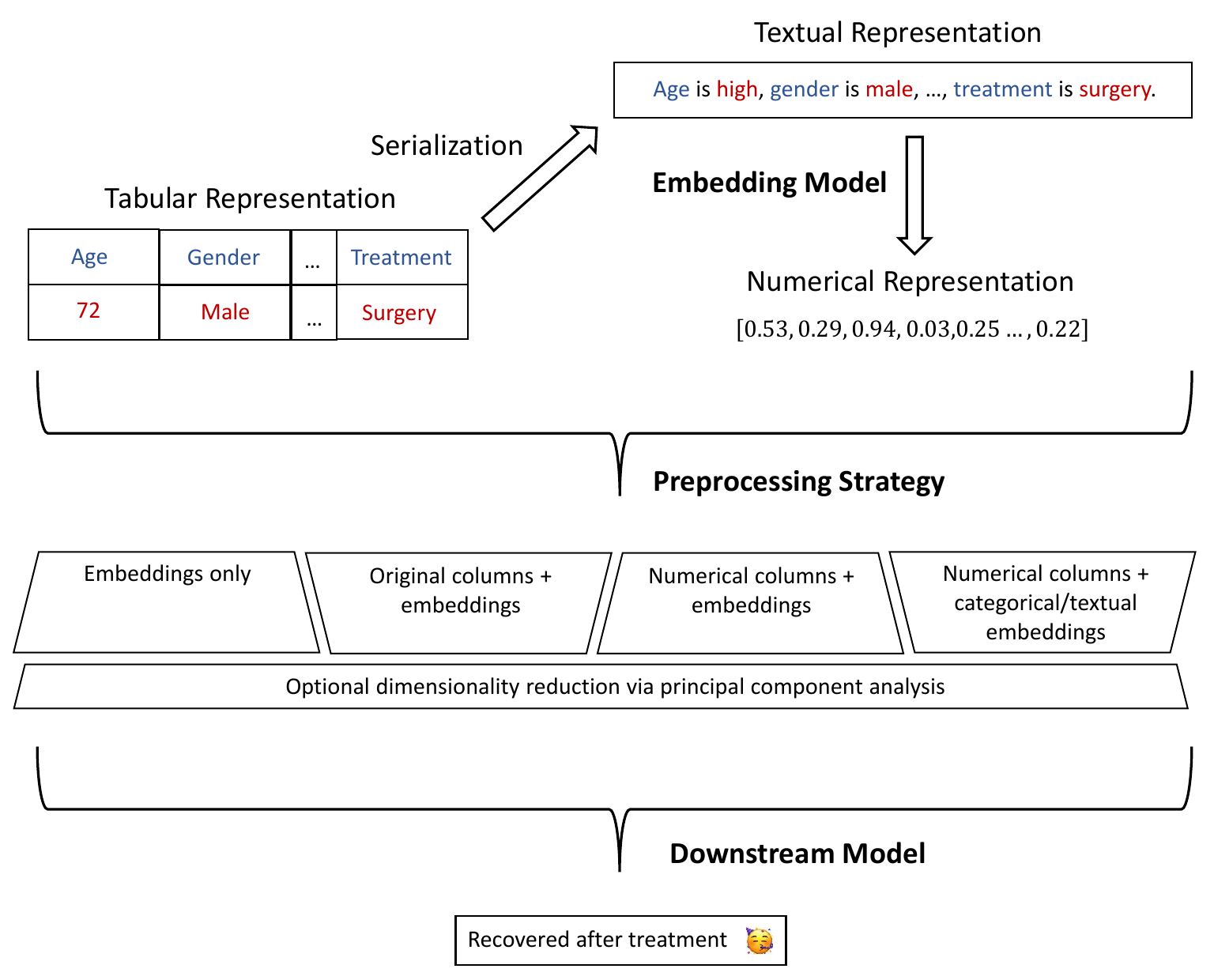}
	\caption{Overview of our work. We serialize tabular rows into text and use embedding models to convert them into numerical representations. These embeddings are preprocessed and serve as inputs for downstream models. We systematically evaluate 256 pipeline configurations, \textit{i.e.}, 16 embedding models, 8 preprocessing strategies, and 2 downstream models.}
	\label{fig:overview}
\end{figure}

Therefore, our paper focuses on systematically benchmarking different preprocessing options, embedding models, and downstream models for tabular prediction. Please refer to Figure~\ref{fig:overview} for an overview of our work. Our contributions are as follows:
\begin{enumerate}
    \item Our empirical results demonstrate that the impact of enriching machine learning models with the LLMs’ world knowledge strongly depends on the specific embedding pipeline configuration. \textbf{We find that concatenating embeddings tends to be better than replacing the original columns with embeddings and that gradient boosting decision trees tend to be strong downstream models} (Sect. Predictive Performance).
    \item We also evaluate the relationship between the predictive performance and embedding model characteristics. \textbf{We show that larger models tend to yield better results, whereas public leaderboard rankings and model popularity are poor performance indicators for tabular prediction tasks} (Sect. Linking Model Properties to Performance).
\end{enumerate}

\section{Related Work}

Carballo et al. (2023) \cite{Carballo2023} were the first to demonstrate that LLM-based embedding approaches can improve the predictive performance on tabular tasks. In their study, they included relevant meta-data, excluded missing values, converted numerical into categorical features, and then adopted a sentence-based serialization - one of the most common representations for tabular prediction \cite{Fang2024}. 

Since then, several authors have explored alternative preprocessing strategies, embedding models, and downstream models. Grinsztajn et al. (2023) \cite{Grinsztajn2023} replaced high-cardinality string features with embeddings and applied dimensionality reduction via principal component analysis (PCA) to them. They tested 26 embedding models, including \texttt{bge-large-en-v1.5} and \texttt{gte-large}, and used scikit-learn's gradient boosting decision tree as a downstream model. Spinaci et al. (2024) \cite{Spinaci2024} used a similar pipeline, but with \texttt{all-MiniLM-L6-v2} as the embedding model and XGBoost or CatBoost as the downstream model. Kasneci and Kasneci (2024) \cite{Kasneci2024} preprocessed the embeddings with PCA and random forest-based feature selection, and assessed 2 preprocessing strategies (replacing versus concatenating the original columns with the embeddings), 3 embeddings configurations (RoBERTa, GPT-2, or both), and 3 downstream models (random forest, XGBoost, or CatBoost). Kim et al. (2024) \cite{Kim2024} skipped PCA altogether and used \texttt{e5-small-v2} and XGBoost with or without bagging. Most recently, Koloski et al. (2025) \cite{Koloski2025} replaced all columns with embeddings, also without PCA. They employed 3 embedding models (\texttt{bge-base-en-v1.5}, unspecified MiniLM, or unspecified Llama 3) in combination with 3 downstream models (multi-layer perceptron, ResNet, or FT-Transformer).

Overall, several studies have explored LLM-based embeddings for tabular prediction, but typically under narrow design choices. As a result, best practices for developing effective embedding pipelines remain unclear. In the following, we systematically benchmark 256 embedding pipeline configurations, \textit{i.e.}, 8 preprocessing strategies, 16 embedding models, and 2 downstream models - much more than any prior work. We then assess whether the predictive performance and embedding model characteristics are sufficiently correlated to guide embedding pipeline design. 

\section{Experimental Setup}

\subsection{Datasets}

To avoid data leakage, we focused our empirical study on datasets that were published no earlier than 2025. Hence, the evaluated embedding models cannot have seen the datasets during their pretraining. We followed the selection criteria of TabArena \cite{Erickson2025} and chose datasets with a mixture of numerical and categorical/textual features from Kaggle:
\begin{itemize}
    \item The lungs diseases dataset \cite{Dalvi2025} is a binary classification dataset that aims to predict whether a patient recovered after treatment or not. It contains 5200 samples, 3 numerical features (age, lung capacity, hospital visits), 2 binary features (gender, smoking status), and 2 textual features (disease type, treatment type). The dataset is almost balanced (2408:2492) and there are 300 missing values per feature. We excluded the 300 missing examples in the prediction target.
    \item The cybersecurity intrusion detection dataset \cite{Samudrala2025} is a binary classification dataset that aims to predict whether or not a cybersecurity attack was detected. It contains 9537 samples, 5 numerical features (network packet size, login attempts, session duration, IP trustworthiness score, failed login attempts), 1 binary feature (unusual time access), and 3 textual features (protocol type, encryption type, browser type). We excluded the session ID feature. The dataset is relatively balanced (4264:5273) and there are no missing values.
\end{itemize}

\subsection{Embedding Pipelines}

\begin{table}[tbp]
  \begin{center}
  \caption{Embedding model characteristics.}
  \begin{tabular}{p{4.4cm}p{1.1cm}p{1.1cm}p{1.1cm}p{0.9cm}}
	\toprule
    Embedding model & Number of para-meters & Embed-ding di-mension & Down-loads August 2025 & Release date\\
    \midrule
    \texttt{all-MiniLM-L6-v2} & 22M & 384 & 89.63M & 2021 \\
    \texttt{bge-base-en-v1.5} & 109M & 768 & 3.32M & 2023 \\
    \texttt{bge-large-en-v1.5} & 335M & 1024 & 3.63M & 2023 \\
    \texttt{bge-small-en-v1.5} & 33M & 384 & 4.35M & 2023 \\
    \texttt{e5-base-v2} & 109M & 768 & 0.88M & 2022 \\
    \texttt{e5-large-v2} & 335M & 1024 & 0.91M & 2022 \\
    \texttt{e5-small-v2} & 33M & 384 & 0.29M & 2022 \\
    \texttt{ember-v1} & 335M & 1024 & 0.04M & 2023 \\
    \texttt{GIST-Embedding-v0} & 335M & 1024 & 0.36M & 2024 \\
    \texttt{GIST-large-Embedding-v0} & 109M & 768 & 0.02M & 2024 \\
    \texttt{GIST-small-Embedding-v0} & 33M & 384 & 0.46M & 2024 \\
    \texttt{gte-base} & 110M & 768 & 0.67M & 2023 \\
    \texttt{gte-base-en-v1.5} & 137M & 768 & 0.27M & 2024 \\
    \texttt{gte-large} & 330M & 1024 & 1.20M & 2023 \\
    \texttt{gte-small} & 30M & 384 & 0.37M & 2023 \\
    \texttt{stella\_en\_400M\_v5} & 435M & 1024 & 1.39M & 2024 \\
    \bottomrule
    \end{tabular}
  \label{tab:characteristics}
  \end{center}
\end{table}

We followed Carballo et al. (2023) \cite{Carballo2023} to serialize our tabular rows into text. We then evaluated 256 pipeline configurations, \textit{i.e.}, 16 embedding models across 2 downstream models and 8 preprocessing strategies. The predictive performance was assessed using a 80\%/20\% train/test split and hyperparameters were selected using a stratified, 5-fold cross-validation and a deviance validation score. All experiments were implemented using scikit-learn 1.5.2 and executed on a machine with an NVIDIA A100 GPU, 16 CPU cores, and 20GB of host memory.

\subsubsection{Embedding Models}

We selected a representative set of 16 LLM-based embedding models from the Massive Text Embedding Benchmark (MTEB) leaderboard \cite{Muenninghoff2023} with $\leq$ 1 billion parameters, including \texttt{all-MiniLM-L6-v2} \cite{Spinaci2024}, \texttt{bge-large-en-v1.5} \cite{Grinsztajn2023}, \texttt{bge-base-en-v1.5} \cite{Koloski2025}, \texttt{e5-small-v2} \cite{Kim2024}, and \texttt{gte-large} \cite{Grinsztajn2023} from prior work. Our models are encoder-only (BERT-style) models with a masked language model pretraining objective. Please refer to Table~\ref{tab:characteristics} for further details. Embeddings were extracted from the \textit{[CLS]} token or via mean pooling that either included or excluded the \textit{[CLS]} and \textit{[SEP]} tokens \cite{Reimers2019}. The extraction strategy was treated as a hyperparameter.

\subsubsection{Downstream Models}

We considered 2 downstream models: L2-regularized logistic regression (LR) and histogram-based gradient boosting decision trees (GBDTs), inspired by LightGBM \cite{Ke2017}. We tuned the L2-regularization strength for LR on the grid [2, 10, 50] and the minimum number of samples per leaf for the GBDTs on the grid [5, 10, 15, 20].

\subsubsection{Preprocessing Strategies}

We adopted 8 preprocessing strategies based on prior work: replacing the original columns with the embeddings \cite{Kasneci2024,Kim2024,Koloski2025}, concatenating the original columns with the embeddings (conc 1) \cite{Kasneci2024}, concatenating the numerical columns with the embeddings (conc 2) \cite{Kim2024}, and concatenating the numerical columns with the embeddings of the categorical/textual columns (conc 3) \cite{Grinsztajn2023,Spinaci2024}. We additionally applied dimensionality reduction via PCA to 50 components \cite{Spinaci2024,Kasneci2024} or no reduction \cite{Kim2024,Koloski2025}. When PCA was used, we normalized our text embeddings prior to reduction; when PCA was not used, we min-max scaled our embeddings for LR.

\subsection{Baselines}

As a non-LLM baseline, we included unsupervised random trees embeddings (RTE) \cite{Moosmann2006}. Here, the maximum tree depth (2 or 5) and number of trees (10 or 100) were set as hyperparameters. Furthermore, both downstream models were run on the raw data as baselines, without any text or random trees embeddings. Binary and textual features were treated as nominal \cite{Kim2024}. For RTE and LR, nominal values were imputed with the mode if missing and one-hot encoded, while missing numerical values were imputed with an iterative imputer (inspired by MICE \cite{vanBuuren2011}). Numerical values were also min-max scaled for LR. Our GBDTs natively support missing values and nominal features, so we only passed nominal feature indicators to them.

\subsection{Evaluation Metrics}

We evaluated the discriminative performance on the test sets using the area under the receiver operating characteristic curve (AUC). In addition, we report the macro-averaged F1-score (F1) and the balanced accuracy (BA) on the test sets.

To examine whether embedding model characteristics can provide guidance for embedding pipeline design, we analyzed the Spearman rank correlation between the predictive performance and several attributes: the MTEB leaderboard score \cite{Muenninghoff2023}, the information sufficiency score \cite{Darrin2024}, the TransformerRanker score \cite{Garbas2025}, the number of parameters, the embedding dimension, the number of downloads in August 2025 (as a proxy for popularity), and the release date (Table~\ref{tab:characteristics}). The information sufficiency and TransformerRanker scores serve as unsupervised metrics to estimate the suitability of embedding models for downstream tasks \cite{Darrin2024,Garbas2025}.

\section{Experimental Results}

\subsection{Predictive Performance}

\begin{figure}[tbp]
	\centering
	\includegraphics[width=\textwidth]{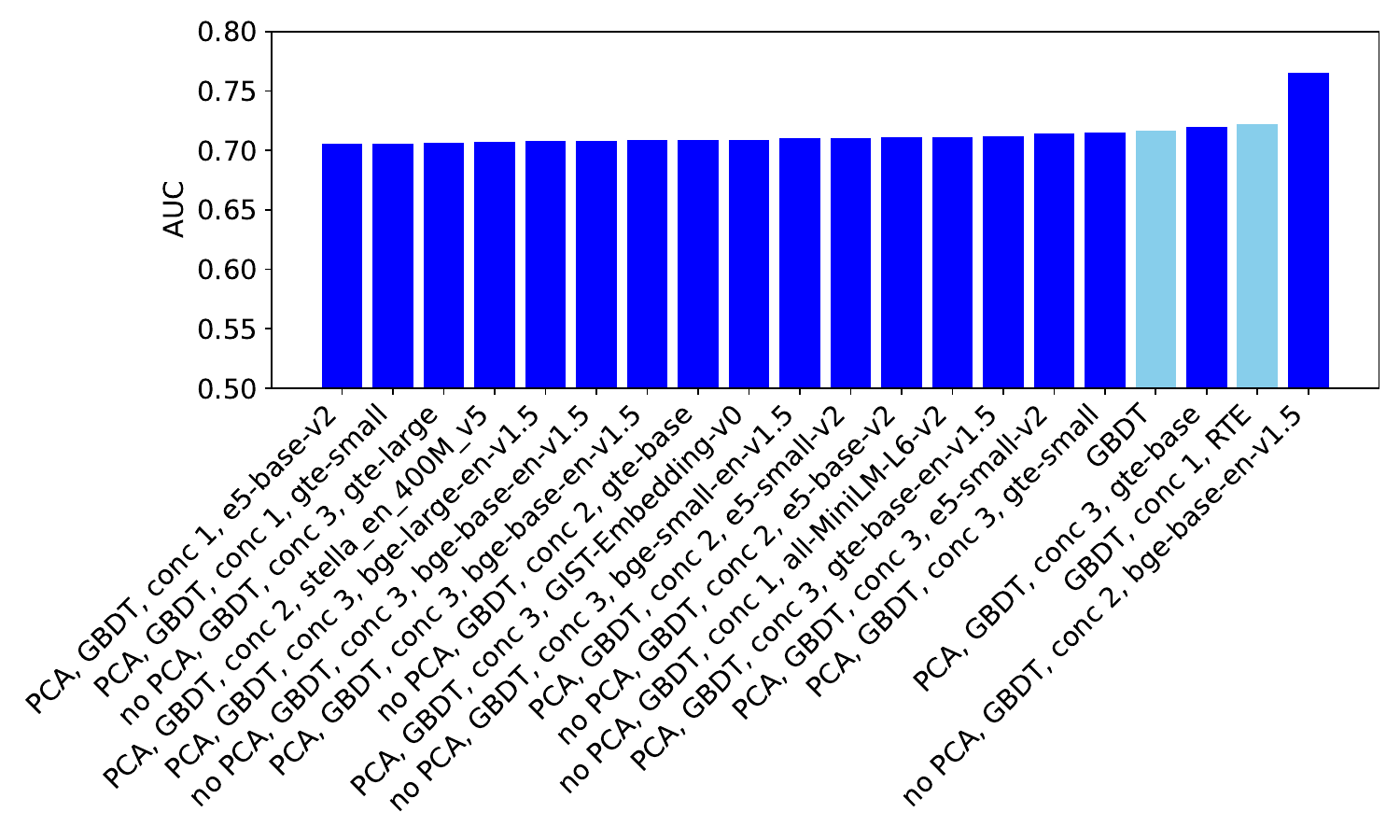}
	\caption{Top-20 mean embedding pipeline performance. Baselines are shown in light blue.}
	\label{fig:top20}
\end{figure}

Out of the 256 embedding pipeline configurations, only 1 pipeline achieved an appreciably better performance than vanilla GBDTs (Figure~\ref{fig:top20} and Table~\ref{tab:performance} in the Appendix). The pipeline employed \texttt{bge-base-en-v1.5} without PCA, concatenated the numerical columns with the embeddings, and used GBDTs as the downstream model. With a mean test AUC of 0.77, it outperformed vanilla GBDTs by 0.05.

In contrast to prior work \cite{Koloski2025}, most pipeline configurations benefited from concatenating embeddings instead of replacing the original columns with embeddings (Figure~\ref{fig:conc_gain} in the Appendix). Applying PCA showed mixed results (Figure~\ref{fig:pca_gain} in the Appendix). Moreover, GBDTs generally achieved a better downstream performance than LR (Figure~\ref{fig:lr_no_pca} to \ref{fig:gbdt_pca} in the Appendix). The best embedding model, on the other hand, could vary significantly depending on the preprocessing strategy and downstream model, highlighting that identifying consistently effective embedding models is challenging (Figure~\ref{fig:lr_no_pca} to \ref{fig:gbdt_pca} in the Appendix).

\subsection{Linking Model Properties to Performance}

\begin{figure}[tbp]
	\centering
	\includegraphics[width=\textwidth]{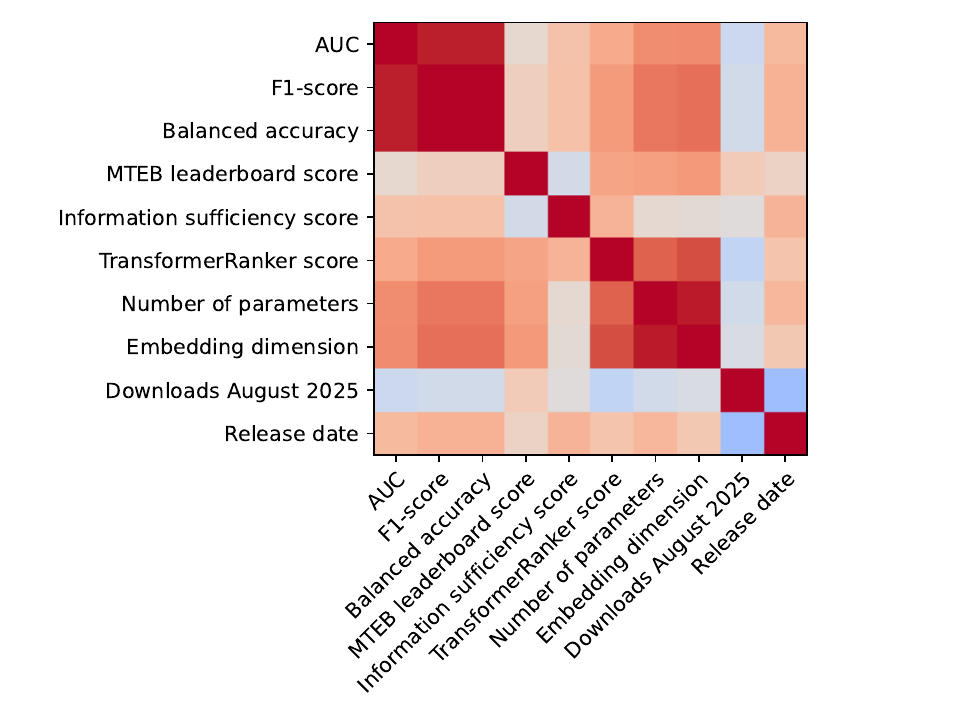}
	\caption{Correlation matrix heatmap between the predictive performance and embedding model attributes, based on the Spearman rank correlation.}
	\label{fig:corr_matrix}
\end{figure}

Embedding model characteristics can provide some guidance to embedding pipeline design (Figure~\ref{fig:corr_matrix}). We observed a moderate correlation (0.56) between the discriminative performance and both the number of parameters and embedding dimension. Our best-performing pipeline configuration was based on \texttt{bge-base-en-v1.5}, though, which only has 109M parameters. The TransformerRanker score (0.40) and the information sufficiency score (0.26) showed weaker associations. Surprisingly, the MTEB leaderboard score was almost uncorrelated with the discriminative performance (0.08), while the number of downloads in August 2025 even showed a slight negative association (-0.12). These findings indicate that larger embedding models are more promising for tabular prediction tasks, whereas public leaderboard rankings and model popularity provide little guidance in selecting effective models.

\section{Conclusion}

Embedding approaches allow researchers and practitioners to efficiently enrich data-driven machine learning methods with the world knowledge of LLMs. Based on a systematic benchmark of 256 pipeline configurations, we show that it strongly depends on the specific pipeline design whether incorporating the prior knowledge of LLMs improves the predictive performance. In general, concatenating embeddings tends to perform better than replacing the original columns with embeddings. While larger embedding models tend to deliver better results, their ranking on public leaderboards and their popularity are poor indicators of performance. Finally, gradient boosting decision trees tend to be strong downstream models. Together, our results offer guidance for building more effective embedding pipelines for tabular prediction and lay the groundwork for future research on compressible embedding models \cite{Choi2025,Nussbaum2025,Zhuang2025}, embedding-specific downstream models \cite{Janson2022}, or alternative serialization options \cite{Fang2024}.

\section*{Acknowledgements}

This research was funded by the Bundesministerium für Bildung und Forschung (BMBF, project number: 16DHBKI071) and the Deutsche Forschungsgemeinschaft (DFG, German Research Foundation, project number: 528483508).

\newpage
\section*{Appendix}

\begin{table}[h]
  \begin{center}
  \caption{Top-20 mean embedding pipeline performance.}
  \begin{tabular}{p{7.0cm}p{0.7cm}p{0.7cm}p{0.7cm}}
	\toprule
    Embedding pipeline & AUC & F1 & BA \\
    \midrule
    no PCA, GBDT, conc 2, \texttt{bge-base-en-v1.5} & 0.77 & 0.76 & 0.75 \\
    GBDT, conc 1, RTE & 0.72 & 0.71 & 0.71 \\
    PCA, GBDT, conc 3, \texttt{gte-base} & 0.72 & 0.71 & 0.71 \\
    GBDT & 0.72 & 0.70 & 0.70 \\
    PCA, GBDT, conc 3, \texttt{gte-small} & 0.72 & 0.70 & 0.70 \\
    PCA, GBDT, conc 3, \texttt{e5-small-v2} & 0.71 & 0.70 & 0.70 \\
    PCA, GBDT, conc 3, \texttt{gte-base-en-v1.5} & 0.71 & 0.70 & 0.70 \\
    no PCA, GBDT, conc 1, \texttt{all-MiniLM-L6-v2} & 0.71 & 0.70 & 0.70 \\
    no PCA, GBDT, conc 2, \texttt{e5-base-v2} & 0.71 & 0.70 & 0.70 \\
    PCA, GBDT, conc 2, \texttt{e5-small-v2} & 0.71 & 0.70 & 0.69 \\
    no PCA, GBDT, conc 3, \texttt{bge-small-en-v1.5} & 0.71 & 0.71 & 0.71 \\
    PCA, GBDT, conc 3, \texttt{GIST-Embedding-v0} & 0.71 & 0.71 & 0.70 \\
    no PCA, GBDT, conc 2, \texttt{gte-base} & 0.71 & 0.70 & 0.69 \\
    PCA, GBDT, conc 3, \texttt{bge-base-en-v1.5} & 0.71 & 0.70 & 0.70 \\
    no PCA, GBDT, conc 3, \texttt{bge-base-en-v1.5} & 0.71 & 0.70 & 0.70 \\
    PCA, GBDT, conc 3, \texttt{bge-large-en-v1.5} & 0.71 & 0.71 & 0.70 \\
    PCA, GBDT, conc 2, \texttt{stella\_en\_400M\_v5} & 0.71 & 0.70 & 0.70 \\
    no PCA, GBDT, conc 3, \texttt{gte-large} & 0.71 & 0.69 & 0.68 \\
    PCA, GBDT, conc 1, \texttt{gte-small} & 0.71 & 0.70 & 0.70 \\
    PCA, GBDT, conc 1, \texttt{e5-base-v2} & 0.71 & 0.70 & 0.70 \\
    \bottomrule
    \end{tabular}
  \label{tab:performance}
  \end{center}
\end{table}

\begin{figure}[tbp]
	\centering
\includegraphics[width=\textwidth]{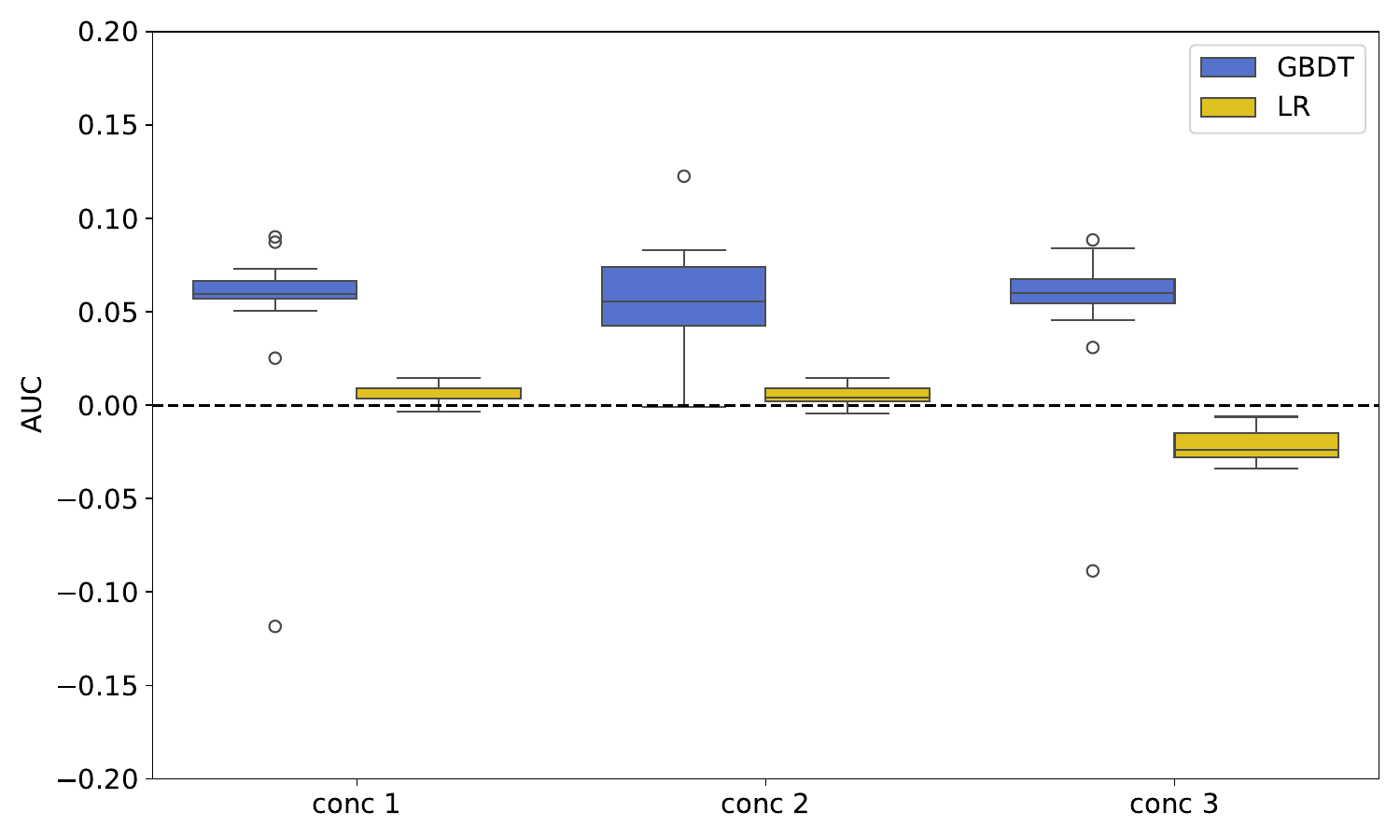}
\caption{Mean AUC gains by concatenating text embeddings instead of replacing the original columns with text embeddings.}
\label{fig:conc_gain}
\end{figure}

\begin{figure}[tbp]
	\centering
\includegraphics[width=\textwidth]{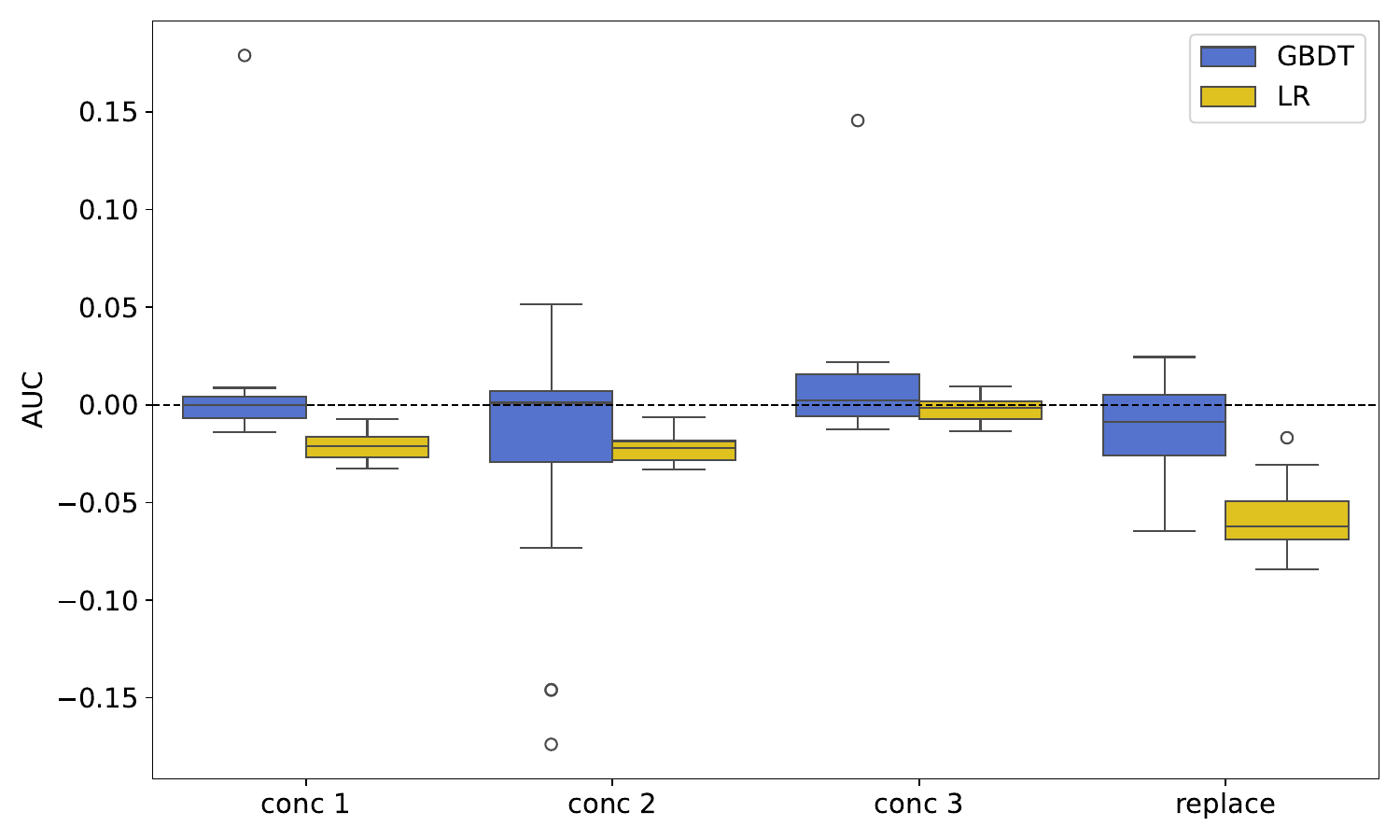}
\caption{Mean AUC gains by applying dimensionality reduction via PCA to text embeddings.}
\label{fig:pca_gain}
\end{figure}

\begin{figure}[tbp]
	\centering
\includegraphics[width=\textwidth]{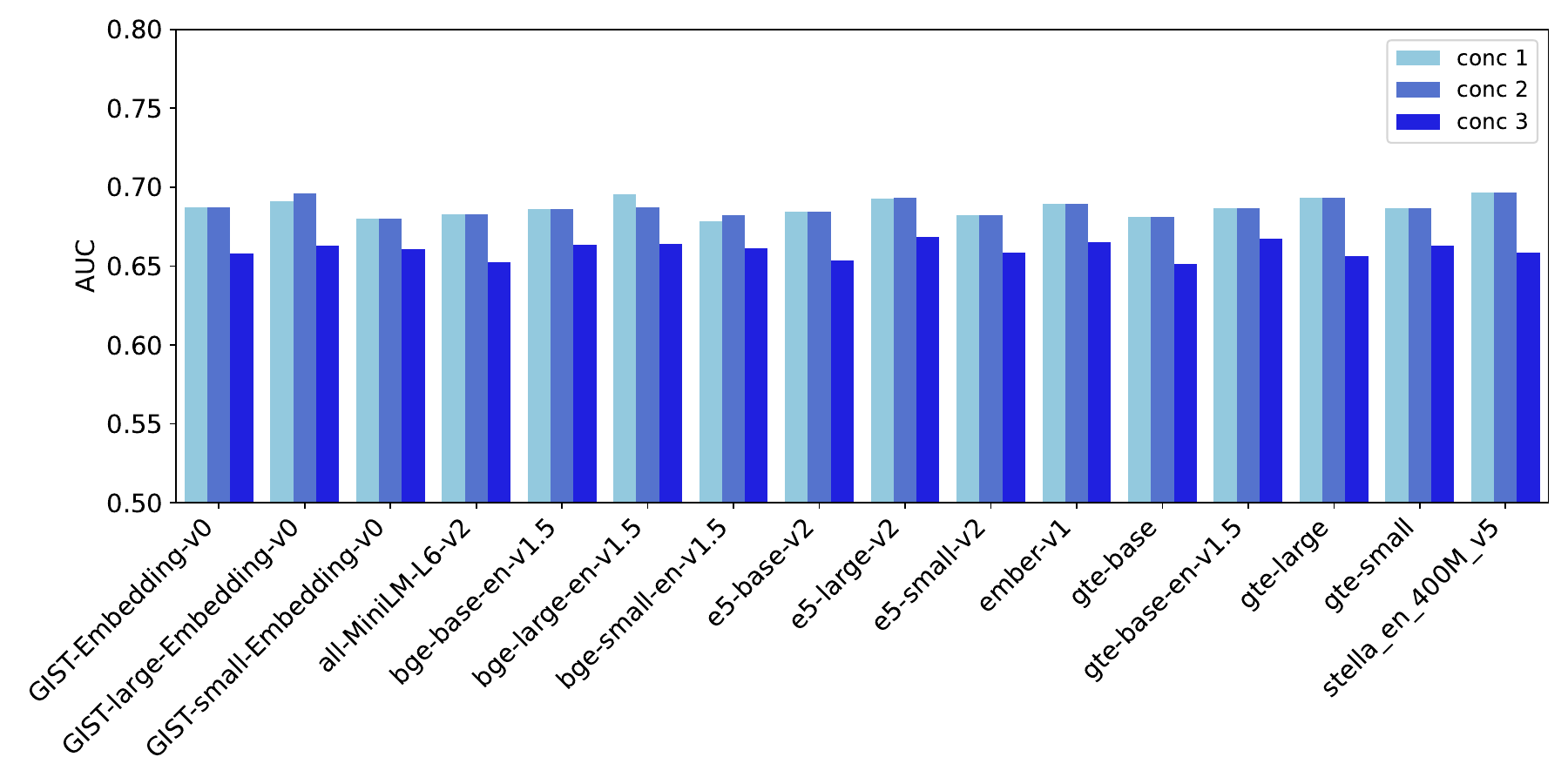}
\caption{Mean embedding model performance for LR without PCA.}
\label{fig:lr_no_pca}
\end{figure}

\begin{figure}[tbp]
	\centering
\includegraphics[width=\textwidth]{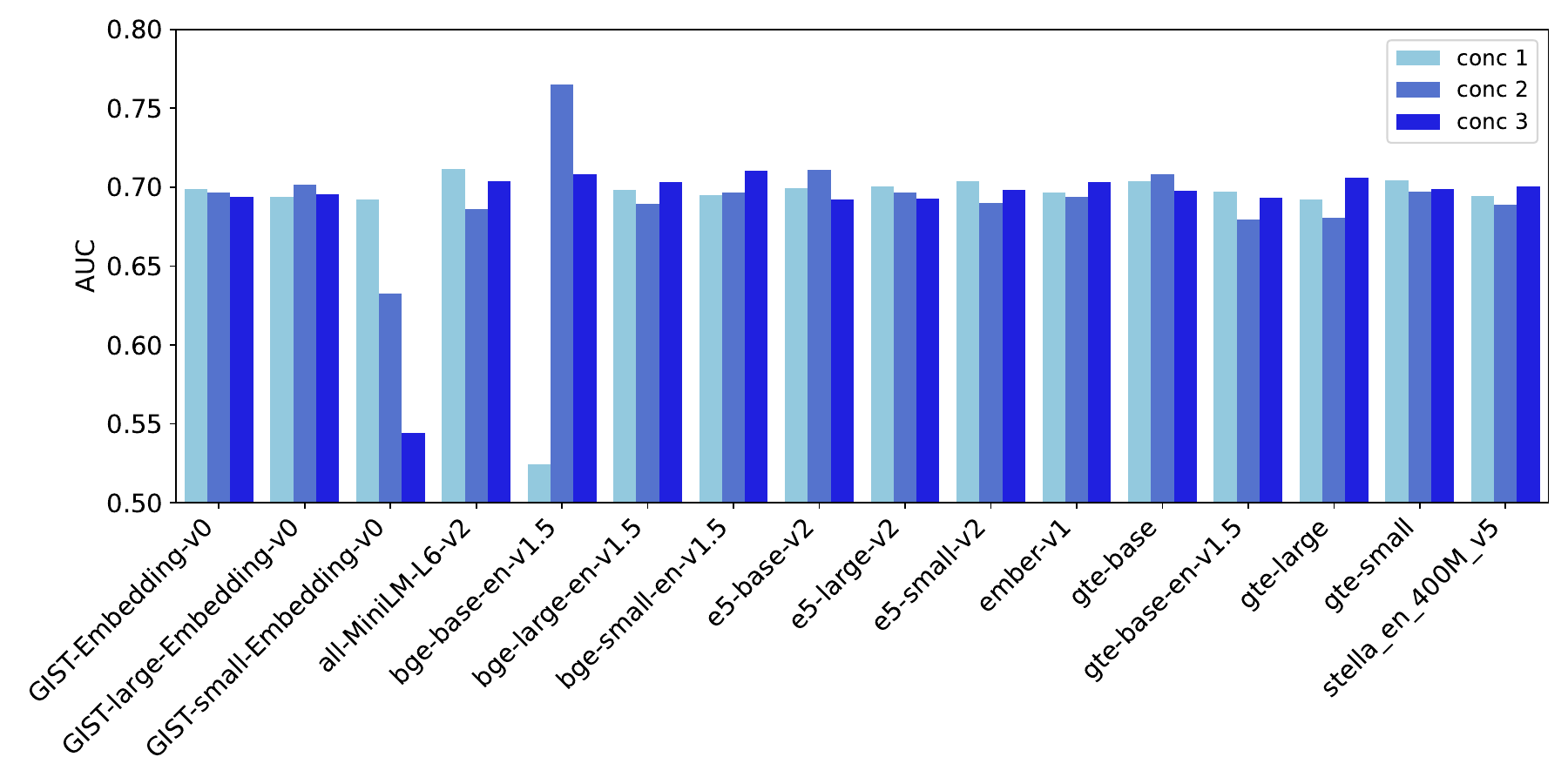}
\caption{Mean embedding model performance for GBDTs without PCA.}
\label{fig:gbdt_no_pca}
\end{figure}

\begin{figure}[tbp]
	\centering
\includegraphics[width=\textwidth]{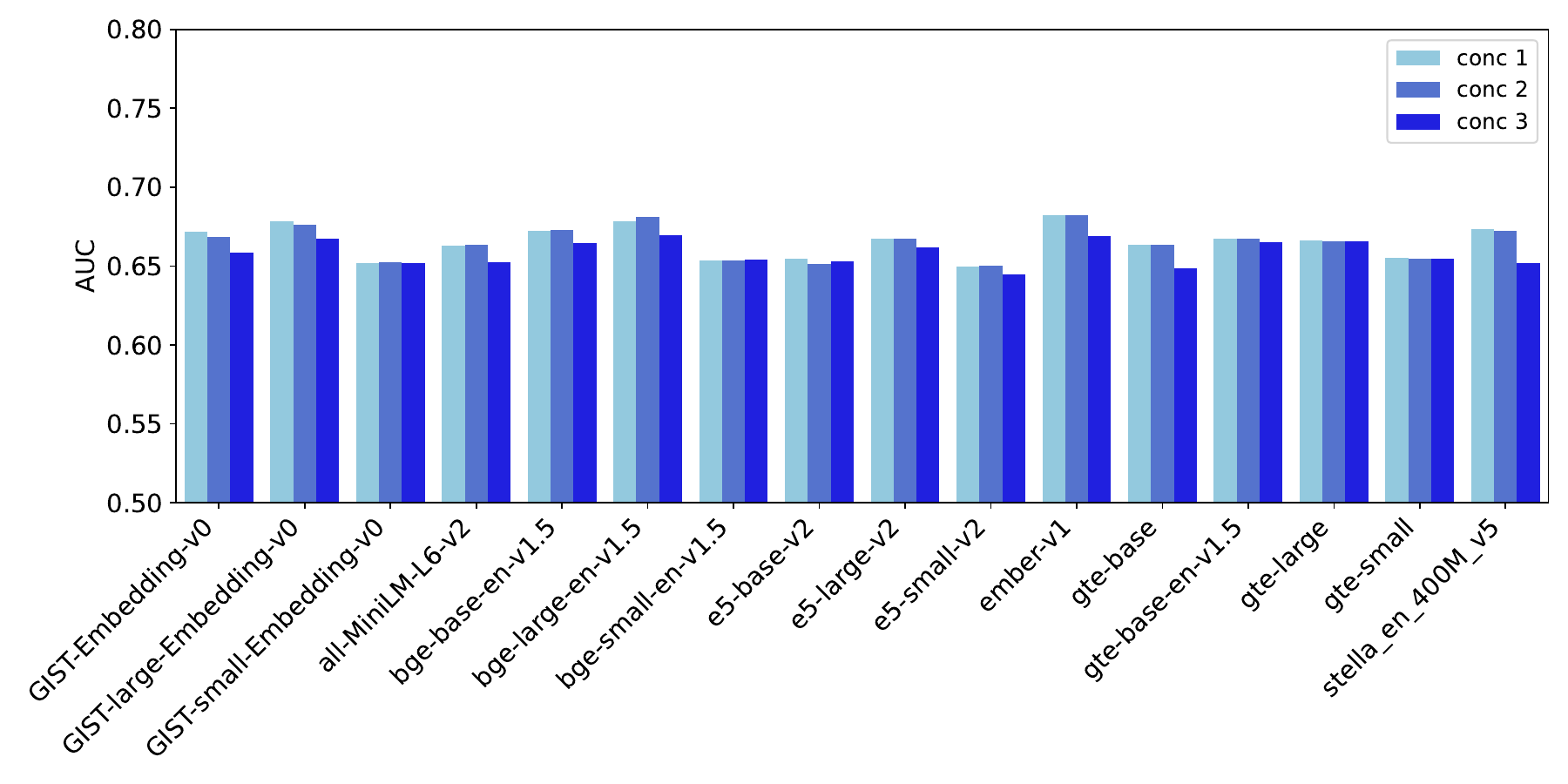}
\caption{Mean embedding model performance for LR with PCA.}
\label{fig:lr_pca}
\end{figure}

\begin{figure}[tbp]
	\centering
\includegraphics[width=\textwidth]{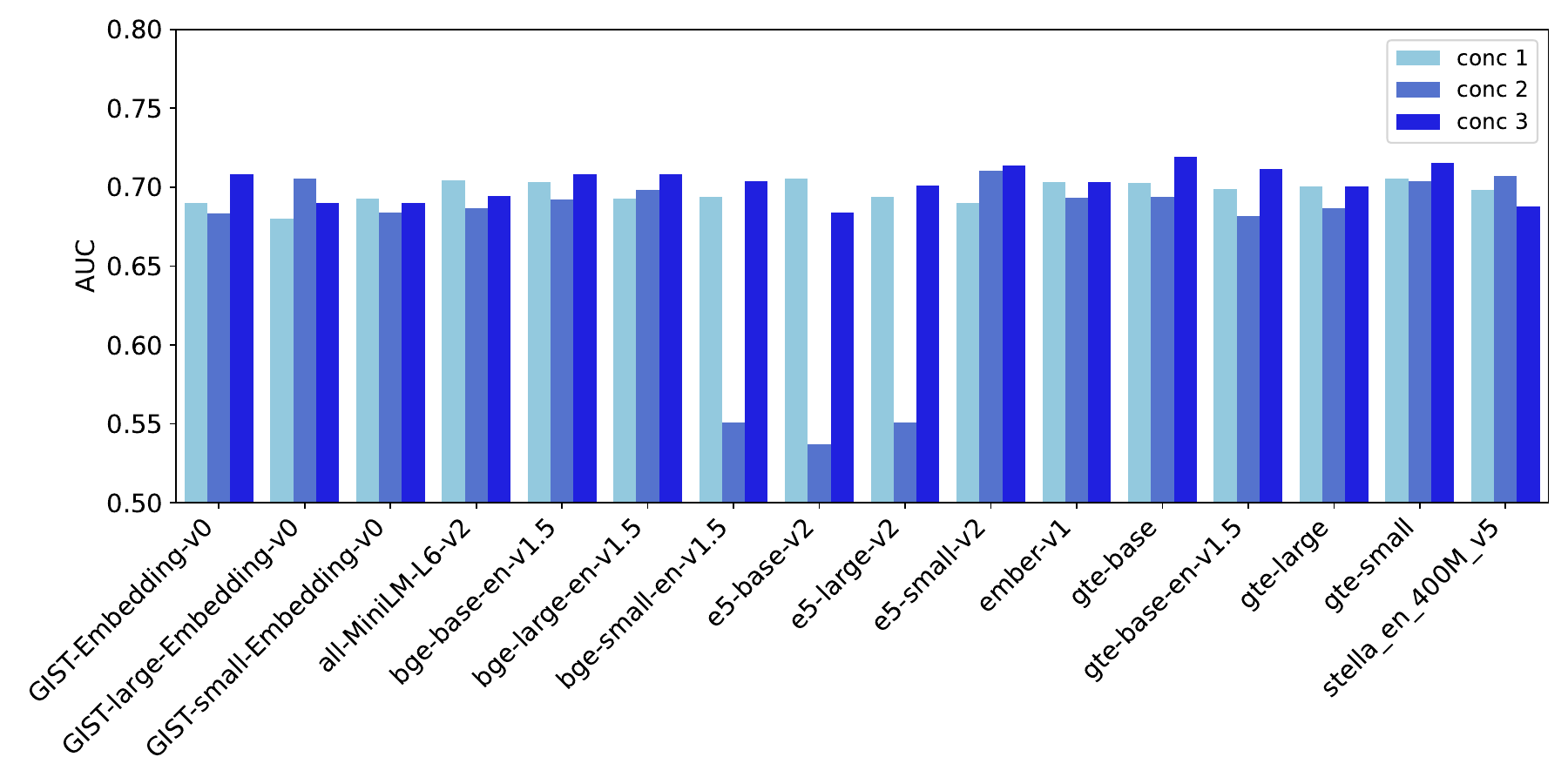}
\caption{Mean embedding model performance for GBDTs with PCA.}
\label{fig:gbdt_pca}
\end{figure}

\addtocontents{toc}{\protect\newpage}



\end{document}